\documentclass{article} 
\usepackage{iclr2024_conference,times}

\usepackage{amsmath,amsfonts,bm}

\def\eqref#1{equation~\ref{#1}}

\def\1{\bm{1}}

\DeclareMathAlphabet{\mathsfit}{\encodingdefault}{\sfdefault}{m}{sl}
\SetMathAlphabet{\mathsfit}{bold}{\encodingdefault}{\sfdefault}{bx}{n}

\usepackage{hyperref}
\usepackage{url}

\title{DragNUWA: Fine-grained Control in Video Generation by Integrating Text, Image, and Trajectory}

 \vspace{-4mm}

\author{\small Shengming Yin$^{1}$\thanks{\scriptsize Both authors contributed equally to this research. Shengming, Jian, and Jie's internship work under the mentorship of Chenfei.} \quad Chenfei Wu$^{2}$\samethanks[1] \quad Jian Liang$^{3}$ \quad  Jie Shi$^{3}$ \quad \textbf{Houqiang Li}$^{1}$\quad \textbf{Gong Ming}$^{2}$\quad \textbf{Nan Duan}$^{2}$\thanks{\scriptsize Corresponding author.} \\
 {$^{1}$University of Science and Technology of China\quad $^{2}$Microsoft Research Asia
 \quad $^{3}$Peking University} \\
 {\tt\scriptsize \{sheyin@mail.,lihq\}@ustc.edu.cn, \{chewu,migon,nanduan\}@microsoft.com, \{j.liang@stu.,jieshi@\}pku.edu.cn }
}

\iclrfinalcopy 

\newcommand*\samethanks[1][\value{footnote}]{\footnotemark[#1]}
\usepackage{graphicx}
\usepackage{floatrow}
\usepackage{tikz}
\usepackage{comment}
\usepackage{amsmath,amssymb} 
\usepackage{color}
\usepackage{graphicx}
\usepackage{amsmath}
\usepackage{amssymb}
\usepackage{multirow}
\usepackage{booktabs}
\usepackage{adjustbox}
\usepackage{siunitx}
\usepackage{graphicx}
\usepackage{caption}
\usepackage{booktabs, makecell}
\usepackage{amsthm}
\usepackage{subcaption}
\usepackage{pgfplots}
\usepackage{multirow}

\begin{document}
\makeatletter
\let\@oldmaketitle\@maketitle
\renewcommand{\@maketitle}{
  \@oldmaketitle
  \vspace{-9mm}
  \begin{minipage}{\linewidth}
    \includegraphics[width=\linewidth]{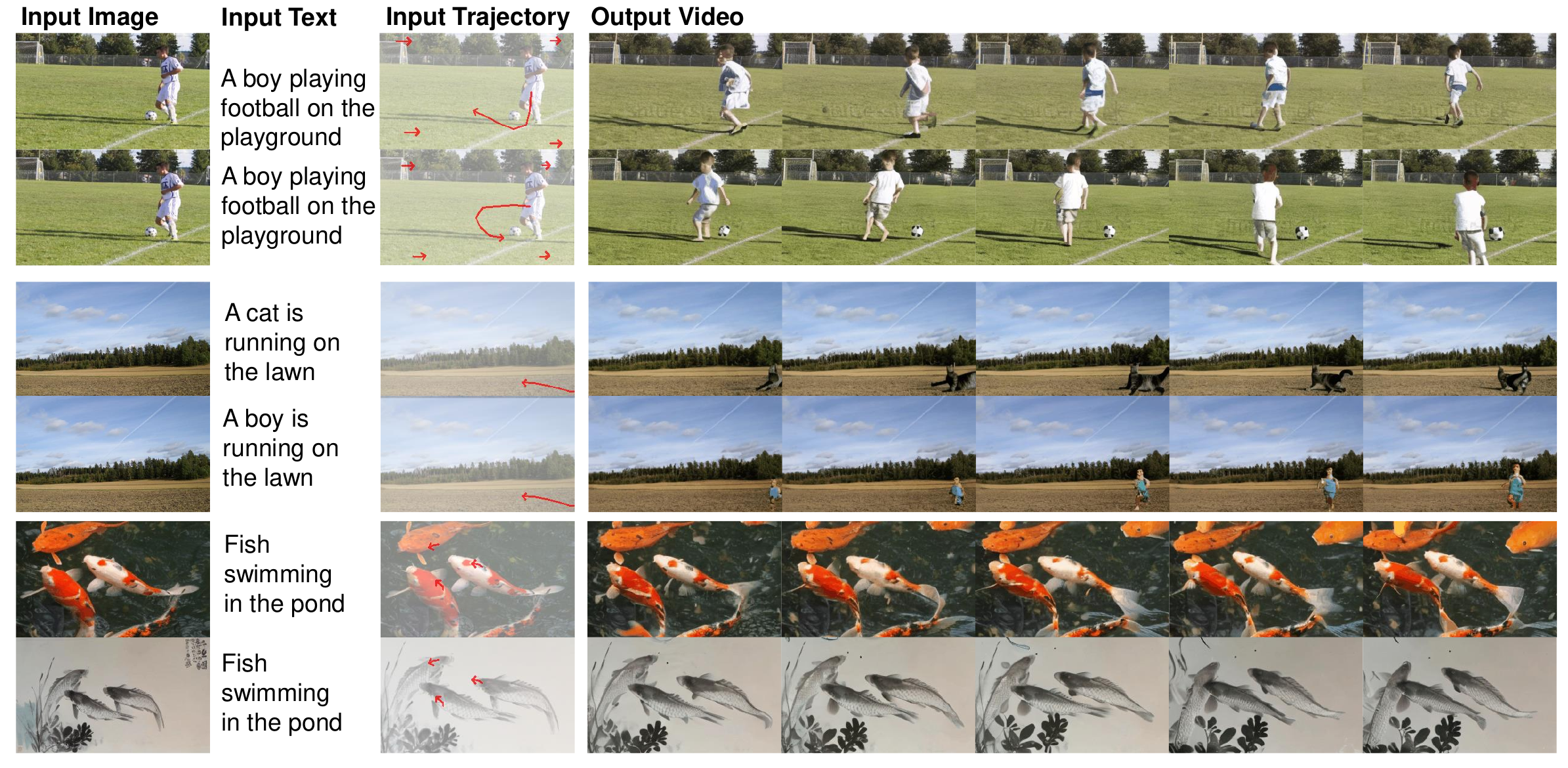} 
    \captionof{figure}{DragNUWA integrates text, image, and trajectory controls to achieve controllable video generation from semantic, spatial, and temporal perspectives. The three groups of examples demonstrate the impact of altering one control while keeping the other two fixed. The first group (Row 1-2) displays the control of complex trajectories, including complex motions (red curved arrows) and camera movements (red rightward arrows). The second group (Row 3-4) illustrates the influence of language control, pairing different text with the same image and trajectory to achieve the effect of introducing new objects in the images. The third group (Row 5-6) demonstrates the impact of image control, showcasing the generation of both real-world and artistic videos.}
    \label{fig:drag}
    \vspace{-2mm}
  \end{minipage}
  }
\makeatother
\maketitle

\begin{abstract}
\vspace{-5mm}

Controllable video generation has gained significant attention in recent years. However, two main limitations persist: Firstly, most existing works focus on either text, image, or trajectory-based control, leading to an inability to achieve fine-grained control in videos. Secondly, trajectory control research is still in its early stages, with most experiments being conducted on simple datasets like Human3.6M. This constraint limits the models' capability to process open-domain images and effectively handle complex curved trajectories. In this paper, we propose DragNUWA, an open-domain diffusion-based video generation model. To tackle the issue of insufficient control granularity in existing works, we simultaneously introduce text, image, and trajectory information to provide fine-grained control over video content from semantic, spatial, and temporal perspectives. To resolve the problem of limited open-domain trajectory control in current research, We propose trajectory modeling with three aspects: a Trajectory Sampler (TS) to enable open-domain control of arbitrary trajectories, a Multiscale Fusion (MF) to control trajectories in different granularities, and an Adaptive Training (AT) strategy to generate consistent videos following trajectories. Our experiments validate the effectiveness of DragNUWA, demonstrating its superior performance in fine-grained control in video generation. The homepage link is \url{https://www.microsoft.com/en-us/research/project/dragnuwa/}

\begin{figure}[p]
    \vspace*{-1cm}
    \makebox[\linewidth]{
        \includegraphics[width=\linewidth]{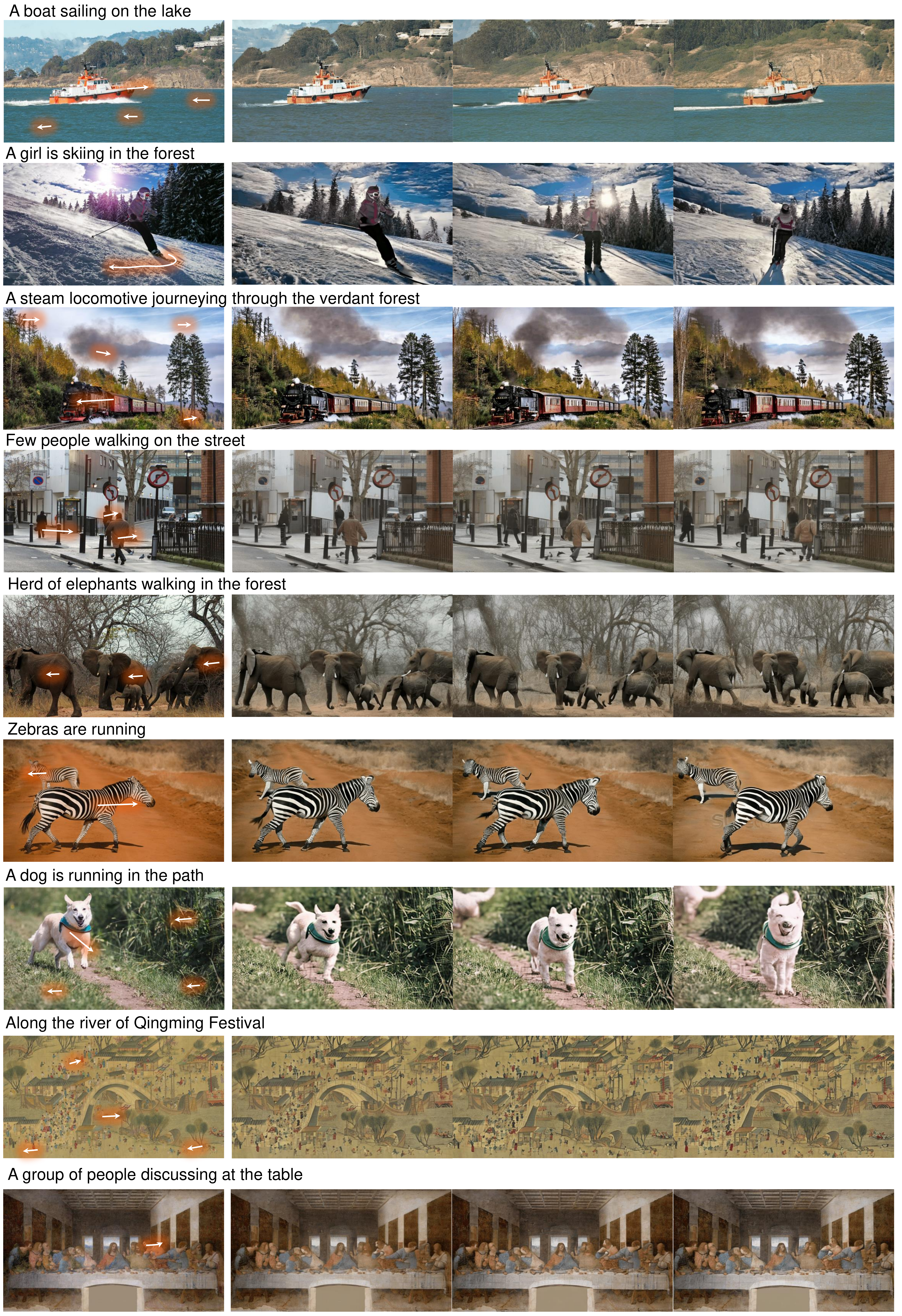}
    }
    \caption{Samples generated by DragNUWA are presented, with the first column showcasing three input controls: text, image, and trajectory. The second, third, and fourth columns exhibit the 5th, 10th, and 15th frames of the output video, respectively. There are 16 frames with a resolution of $576\times 320$ in total.  DragNUWA is capable of concurrently controlling the movement of the camera, multiple objects, and complex trajectories, enabling the generation of videos featuring both real-world scenes and artistic paintings.}
\end{figure}

\end{abstract}

\section{Introduction}

Controllable video generation is a hot topic in research. Most of these studies focus on controllable visual generation. Early research primarily emphasized image-to-video generation, using the initial frame image as a control to manipulate the generated video spatially~\citet{lotterDeepPredictiveCoding2016,srivastavaUnsupervisedLearningVideo2015,chiappaRecurrentEnvironmentSimulators2016}. However, relying solely on images as controls cannot determine the subsequent frames of future videos. Consequently, there has been growing interest in text-to-video research, employing text to semantically constrain video generation~\citet{wuGODIVAGeneratingOpenDomaIn2021,wuUWAVisualSynthesis2022,hongCogVideoLargescalePretraining2022,singerMakeAVideoTexttoVideoGeneration2022,hoImagenVideoHigh2022}. Some studies also utilize both text and image conditions for more precise control over video generation~\citet{huMakeItMove2022,yinNUWAXLDiffusionDiffusion2023,esserStructureContentguidedVideo2023}. Nonetheless, both language and image remain limited in expressing the temporal information of videos, such as camera movements and complex object trajectories.

To control temporal information of videos, trajectory-based control has emerged as a user-friendly approach increasingly gaining attention in research. CVG~\citet{haoControllableVideoGeneration2018} and C2M~\citet{ardinoClickMoveControlling2021} encode images and trajectories, predicting optical flow maps and warp features as intermediate results for controllable video generation. However, warp operations often result in unnatural distortions. To solve this issue, II2V~\citet{blattmannUnderstandingObjectDynamics2021} and iPOKE~\citet{blattmannIpokePokingStill2021} compress videos into a dense latent space and learn to manipulate these latent variables using RNN. Similarly, MCDiff~\citet{chenMotionConditionedDiffusionModel2023} predicts future frames by diffusion latents in an auto-regressive way. While  MCDiff has achieved promising results, it relies on HRNet~\citet{wangDeepHighResolutionRepresentation2021} to extract 17 keypoints for each person to construct data, it can only control motion from humans. Additionally, MCDiff and the aforementioned models neglect to consider the control of languages, which in turn limits their ability to control the videos effectively.

The aforementioned research inspired us with a two-fold vision for controllable video generation. 1) Firstly, the current consideration of text, image, and trajectory-based controls in existing works is not comprehensive enough. We argue that these three types of controls are indispensable, as they each contribute to the regulation of video content from semantic, spatial, and temporal perspectives. As depicted in Figure~\ref{fig:drag}, the combination of text and images alone is insufficient to convey the intricate motion details present in a video, which can be supplemented by incorporating trajectory information. Furthermore, while images and trajectories may not adequately represent future objects in a video, language can compensate for this shortcoming. Lastly, relying solely on trajectories and language can result in ambiguity when expressing abstract concepts, such as differentiating between real-world fish and a painting of a fish, whereas images can provide the necessary distinction. 2) Secondly, current research on trajectory control is still in its early stages, with most experiments being conducted on simple datasets like Human3.6M. This constraint limits the models' capability to process open-domain images and effectively handle complex curved trajectories, multiple object movements, and camera motion simultaneously.

Based on these observations, we propose DragNUWA, an open-domain video generation model. To address the issue of insufficient control granularity in existing works, we simultaneously introduce text, image, and trajectory information to provide fine-grained control over video content from semantic, spatial, and temporal perspectives. To resolve the problem of limited open-domain trajectory control in current research, We model trajectory with three aspects: a Trajectory Sampler (TS) to enable open-domain control of arbitrary trajectories, a Multiscale Fusion (MF) to control trajectories in different granularities, and an Adaptive Training (AT) strategy to generate consistent videos following trajectories.

The main contributions of our work are as follows:

\begin{itemize}
    \item We introduce DragNUWA, an end-to-end video generation model that seamlessly integrates three essential controls—Text, Image, and Trajectory—providing strong and user-friendly controllability.
    \item We focus on trajectory modeling with three aspects: a Trajectory Sampler (TS) to enable open-domain control of arbitrary trajectories, a Multiscale Fusion (MF) to control trajectories in different granularities, and an Adaptive Training (AT) strategy to generate consistent videos following trajectories.
    \item We conduct extensive experiments to validate the effectiveness of DragNUWA, demonstrating its superior performance in fine-grained control in video synthesis.

\end{itemize}

\section{Related Works}

\subsection{Text/Image Control in Video Synthesis}

Early research primarily emphasized image-to-video generation, with a common assumption that the environment is deterministic and has only one possible future~\citet{lotterDeepPredictiveCoding2016,srivastavaUnsupervisedLearningVideo2015,chiappaRecurrentEnvironmentSimulators2016}. However, this assumption cannot satisfy the requirements of real-world videos with unlimited possibilities. To address this issue, text-to-video generation has been widely studied in recent years (GODIVA~\citet{wuGODIVAGeneratingOpenDomaIn2021}, NUWA~\citet{wuUWAVisualSynthesis2022}, CogVideo~\citet{hongCogVideoLargescalePretraining2022}, Make A Video~\citet{singerMakeAVideoTexttoVideoGeneration2022}, Imagen Video~\citet{hoImagenVideoHigh2022}), introducing text descriptions to semantically control the content of video generation. However, text alone cannot accurately describe the spatial information of visuals. Therefore, MAGE~\citet{huMakeItMove2022} emphasizes text-image-to-video, utilizing both semantic information from text and spatial information from images for precise video control. Similarly, GEN-1~\citet{esserStructureContentguidedVideo2023} integrates depth maps with texts using cross-attention mechanisms for control. In the domain of long video generation, text-image-to-video has also been widely used. For example, Phenaki~\citet{villegasPhenakiVariableLength2022} generates subsequent frames by auto-regressively introducing previous frames and text, achieving long video generation. NUWA-XL~\citet{yinNUWAXLDiffusionDiffusion2023} employs a hierarchical diffusion architecture to continuously complete intermediate frames based on previous frames and text.

While text and images can effectively convey semantics and appearance, they struggle to adequately represent complex motion information and camera movements.  Unlike these approaches, DragNUWA adds trajectory control to text and image control, enabling fine-grained control of videos in terms of semantics, appearance, and motion.

\subsection{Trajectory Control in Video Synthesis}

To better control motion in videos, future video prediction methods control subsequent frame generation based on given video frames~\citet{wuFutureVideoSynthesis2020,wichersHierarchicalLongtermVideo2018,walkerPoseKnowsVideo2017,liangDualMotionGAN2017}. On the other hand, video-to-video generation transfers the style of a complete video or video sketch to a new domain, providing rich control information~\citet{chanEverybodyDanceNow2019,wangFewshotVideotovideoSynthesis2019}. However, this requires users to provide video input and restricts fine-grained control as the style transfer is based on the original video's skeleton. Consequently, image-trajectory-to-video methods have emerged, controlling video development through trajectories given in images. CVG~\citet{haoControllableVideoGeneration2018} and C2M~\citet{ardinoClickMoveControlling2021} encode images and trajectories, predicting optical flow maps and warp features as intermediate results for controllable video generation. However, warp operations often result in unnatural distortions. II2V~\citet{blattmannUnderstandingObjectDynamics2021} and iPOKE~\citet{blattmannIpokePokingStill2021} compress videos into a dense latent space and learn to manipulate these latent variables using recurrent neural networks. However, since trajectory control operates on the pixel level, it is sparse and prone to ambiguity. To address this issue, sparse strokes are first transformed into dense flows, and then future frames are predicted based on dense flow using autoregression. Nonetheless, since MCDiff~\citet{chenMotionConditionedDiffusionModel2023} relies on HRNet~\citet{wangDeepHighResolutionRepresentation2021} to extract 17 keypoints for each person to construct data, it can only control motion from humans. To achieve control of open-domain objects, Video Composer~\citet{wangVideoComposerCompositionalVideo2023} very recently used MPEG-4 to extract motion vector information from videos as conditions for training, but due to the lack of high-level semantic information in motion vectors, it could only control simple object movements.

In comparison to previous research, which solely focused on managing human motion or rudimentary object movements, DragNUWA stands out as the pioneering approach in accomplishing fine-grained open-domain video generation by enabling the dragging of any objects in an image, facilitating control over multiple objects, and accommodating their complex trajectories and camera movements.

\begin{figure}[!ht]
    \centering
    \includegraphics[width=\linewidth]{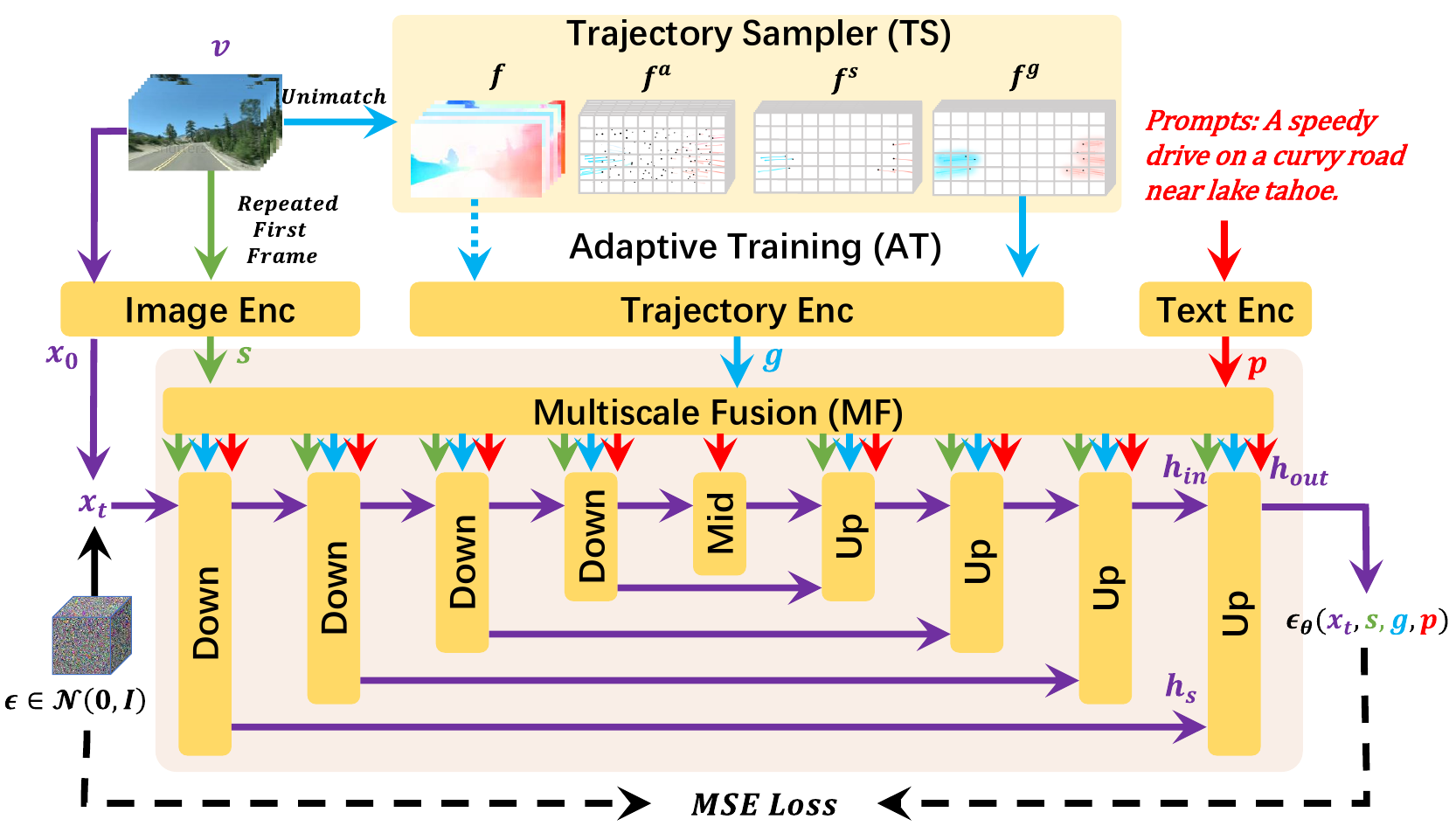}
    \caption{Overview of DragNUWA's Training Process. DragNUWA supports three optional inputs: text $p$, image $s$, and trajectory $g$, and focuses on designing the trajectory from three aspects. First, the Trajectory Sampler (TS) dynamically samples trajectories from open-domain video flow. Second, Multiscale Fusion (MF) deeply integrates trajectory with text and image within each block of the UNet architecture. Lastly, Adaptive Training (AT) adapts the model from optical flow conditions to user-friendly trajectories. Ultimately, DragNUWA is capable of handling open-domain videos with multiple objects and their complex trajectories.}
    \label{fig:training}
\end{figure}

\section{Method}
Unlike previous works that only support either text-based~\citet{wuGODIVAGeneratingOpenDomaIn2021}, image-based~\citet{zhangDTVNetDynamicTimeLapse2020}, or trajectory-based control~\citet{huMakeItMove2022}, DragNUWA is designed to incorporate all three types of control while emphasizing trajectory modeling from three aspects: 

\begin{itemize}
    \item 1) To enable open-domain control of arbitrary trajectories, a Trajectory Sampler (TS) (introduced in Sec.~\ref{sec:ts}) is employed to directly sample trajectories from open-domain video flows during training, as opposed to the specific domain, such as human pose trajectories used in MCDiff~\citet{chenMotionConditionedDiffusionModel2023}. 
    \item 2) To achieve control over different trajectory granularities, a Multiscale Fusion (MF) (introduced in Sec.~\ref{sec:mf}) is utilized to downsample the trajectory to various scales and deeply integrate it with text and image within each block of the UNet architecture, rather than directly concatenating controls with diffusion noise as in~\citet{chenMotionConditionedDiffusionModel2023,wangVideoComposerCompositionalVideo2023}.
    \item 3) To generate stable and consistent videos, we adopt an Adaptive Training (AT) (introduced in Sec.~\ref{sec:at}) approach, initially conditioning on dense flow to stabilize video generation and subsequently training on sparse trajectories to adapt the model.
\end{itemize}

In the following sections, we focus on introducing the training process from Sec.~\ref{sec:ts} to Sec.~\ref{sec:at}, specifically discussing how the model calculates loss by merely using the input video and text pair, denoted as $<v, p>$. In Sec.~\ref{sec:ui}, we introduce the inference process, which demonstrates how the model processes the input text $p$, image $s$, and trajectory $g$ to output the generated video $v$.

\subsection{Trajectory Sampler (TS)}\label{sec:ts}

In the training data,  since it only contains video and text pairs $<v,p>$, it is essential to extract trajectories from the videos. Previous studies primarily utilized key point tracking models to pre-extract video trajectories for training. However, this approach has two main drawbacks. Firstly, as these models are trained on specific domains, such as human poses, their ability to handle open-domain videos is limited. Secondly, in practical applications, it is difficult to ensure that users input trajectories precisely at key points, resulting in a gap between training and inference. To facilitate open-domain video trajectories and enable users to input arbitrary trajectories, We designed a Trajectory Sampler (TS) that directly samples trajectories from video optical flow, allowing the model to learn various possible trajectories in an open-domain setting.

Given a video $v \in \mathbb{R}^{L\times C\times H\times W}$ with $L$ frames, $C$ channels, $H$ height, $W$ width,  we first utilize Unimatch~\citet{xu2023unifying}, an optical flow estimator, to extract dense optical flow $f \in \mathbb{R}^{(L-1)\times C\times H\times W}$. For clarity, we represent the optical flow of the first and second frames as $f_0 \in \mathbb{R}^{C\times H\times W}$. A straightforward approach is to directly sample trajectories from $f_0$ according to the intensity of the optical flow. However, this would result in excessive sampling on objects with larger motions, while those objects with smaller motions would have limited opportunities for learning. To handle this issue, we uniformly distribute anchor points with an interval of $\lambda$. Moreover, to cover the entire image region as much as possible, we add random perturbations $\delta$ ranging from $-\lambda/2$ to $\lambda/2$ to the anchor points. Finally, we obtain a slightly sparser anchored optical flow $f_0^a$ in the following:
\begin{align}\label{eq:fa}
    f_{0, i, j}^a = \begin{cases}
        0, & \texttt{else} \\
        f_{0, i, j}, & (i+\delta)\%\lambda=0 \, \texttt{\&} \, (j+\delta)\%\lambda=0
    \end{cases}
\end{align}
To support control over multiple trajectories, we define the maximum number of trajectories $N$ and randomly sample the number of trajectories $n\sim U[1, N]$. To accommodate both large and small motion objects while selecting trajectories based on flow intensity, we sample $n$ anchor tracking points from $f_{0,i,j}^{a}$ according to the multinomial distribution $M(n, ||f_{0,i,j}^{a}||_2)$. This results in a sparser flow of $f_0^s$ containing $n$ tracking points.
Since $f_0^s$ only contains the tracking points from the first frame, to obtain the full trajectory $f^s$, we proceed to iteratively track the trajectories by updating the position of the tracking points according to the corresponding optical flow $f$.

Given that $f^s$ is highly sparse, it is not conducive for the model to learn from these trajectories. Therefore, we apply Gaussian Filter to $f^s$ to obtain an enhanced trajectory map $f^g\in\mathbb{R}^{(L-1)\times C\times H \times W}$. Compared with  $f^s$, $f^g$ improves the robustness and helps the model to better capture trajectory information.

\subsection{Multiscale Fusion (MF)}\label{sec:mf}
\paragraph{Encoding of Video}
During training, we treat video $v\in\mathbb{R}^{L\times C\times H \times W}$ as independent frames and encode it into $x_0\in \mathbb{R}^{L\times c\times h\times w}$ using a pre-trained image autoencoder~\citet{rombachHighResolutionImageSynthesis2022}. It is important to note that the subscript 0 in $x_0$ does not represent the first frame  but indicates the initial step in the diffusion process. We follow the pre-defined diffusion process $q\left(x_t\middle| x_{t-1}\right)=\mathcal{N}\left(x_t;\sqrt{\alpha_t}\ x_{t-1},\ \left(1-\alpha_t\right)\mathbf{I}\right) $  and add noise to $x_0$:
\begin{align}
    x_t=\sqrt{{\bar{\alpha}}_t}\ x_0+\sqrt{(1-{\bar{\alpha}}_t)}\epsilon\quad \epsilon\sim\mathcal{N}(\mathbf{0},\mathbf{I})
\end{align}
where $\epsilon$ is noise, $x_t$ is the $t$-th intermediate state in diffusion process, $\alpha_t, {\bar{\alpha}}_t$ is hyperparameters in diffusion model.

\paragraph{Encoding of Text Control}
Given the text prompts, we encode them with CLIP~\citet{radfordLearningTransferableVisual2021a} Text Encoder to get prompt embedding $p\in \mathbb{R}^{l_p\times c_p}$ where $l_p$ is token length, $c_p$ is prompt embedding dimension. 

\paragraph{Encoding of Image Control}
For image control, we utilize the first frame of the video as a condition, providing general information such as appearance, style, and layout. To match the size for fusion, the first frame is repeated $L$ times. Subsequently, the pre-trained image autoencoder~\citet{rombachHighResolutionImageSynthesis2022} and a sequence of convolution layers are employed to independently encode each frame into a representation $s \in \mathbb{R}^{L\times c_s\times h\times w}$.

\paragraph{Encoding of Trajectory Control}
By Trajectory Sampler introduced in Sec.~\ref{sec:ts}, we obtain $f^g\in\mathbb{R}^{(L-1)\times C\times H \times W}$ directly from open-domain videos $v$. To match the fusion size, we pad a full zero frame in front of $f^g$, and encode it using a series of convolutional layers, resulting in $g\in\mathbb{R}^{L\times c_g\times h \times w}$. 

To fuse multiple controls in different granularities, we propose Multiscale Fusion (MF), which can simultaneously accept text $p$, image $s$, and trajectory $g$ as conditions and merge them at different resolutions. The Multiscale Fusion will first downsample the trajectory $g$ and image $s$ to various scales $g^{(l)}$ and $s^{(l)}$, where the superscript $l$ represents  downsample depth. The trajectory and image are then integrated with text $p$ in UNet architecture, composed of multiscale downblocks and upblocks with skip connection. 

For the image condition $s$ and trajectory condition $g$, they are fused into hidden state $h$ via linear projection. 
In the $l$-th block of UNet architecture, $s^{(l)}$, $m^{(l)}$ and $g^{(l)}$ are first transferred to scale $w_s^{(l)},w_m^{(l)},w_g^{(l)}$ and shift $b_s^{(l)},b_m^{(l)},b_g^{(l)}$ via zero-initialized convolution layers, where $m^{(l)}$ is a binary mask to indicate whether the frame is provided as a condition. Then, the scale $w$ and shift $b$ are fused into $h$ via simple linear projection.
\begin{align}
    h&:=w_s^{(l)}\cdot h+b_s^{(l)}+h \\
    h&:=w_m^{(l)}\cdot h+b_m^{(l)}+h \\
    h&:=w_g^{(l)}\cdot h+b_g^{(l)}+h 
\end{align}

For the text condition $p$, it is injected to hidden states $h$ via Prompt Cross-Attention with hidden states $h$ treated as query, and text $p$ treated as key and value.

To support various combinations of conditions, we introduce randomness into the training process by randomly omitting text, images, and trajectories before feeding them into Multiscale Fusion. For the dropped text, we employ empty strings as replacements, whereas for dropped images and trajectories, frames populated with all zeros are used. Through this training paradigm involving mixed conditions, our model is capable of generating consistent videos during inference across different condition combinations.

\subsection{Adaptive Training (AT)}\label{sec:at}

Simultaneously conditioning the video generation process on both image and sparse trajectory while maintaining visual consistency presents a significant challenge. To address this issue, we employ an Adaptive Training (AT) strategy to optimize DragNUWA. 

In the first stage, to generate visually and dynamically consistent videos,  we provide the model with prompt $p$, dense optical flow $f$, and the repeated first frame $s$ as conditions, the model is optimized to minimize the distance between the output of the UNet $\epsilon_\theta\left(x_t,p,s,f\right)$ and the added noise $\epsilon$. Considering the density of optical flow, we do not apply Gaussian filtering for enhancement.
\begin{align}
    \mathcal{L}_\theta=\left|\left|\epsilon-\epsilon_\theta\left(x_t,p,s,f\right)\right|\right|_2^2
\end{align}
As provided the complete optical flow $f$ as a condition, it is much easier to generate dynamically consistent videos while preserving the first frame. In the second stage, to adapt the model from complete optical flow to user-friendly trajectories, we continue training the model by sampling trajectory $f_g$ from the original optical flow $f$ using Trajectory Sampler (TS).
\begin{align}
    \mathcal{L}_\theta=\left|\left|\epsilon-\epsilon_\theta\left(x_t,p,s,g\right)\right|\right|_2^2
\end{align}
Despite the trajectory being considerably sparser than the optical flow, the model is capable of generating dynamics consistent with trajectories while maintaining stability and consistency learned from the previous training. 

\subsection{Inference}\label{sec:ui}
During inference, given the text, image, and trajectory, DragNUWA is capable of generating realistic and contextually consistent videos $v$.

The text is encoded by CLIP~\citet{radfordLearningTransferableVisual2021a} Text encoder to get text embedding $p$. The image is repeated $L$ times and encoded to $s$. The input trajectory is first processed by Gaussian Filter and zero frame padding and then encoded to $g$. After that, $x_0$ is iteratively sampled from a pure Gaussian noise $x_T$ using the Unet $\epsilon_\theta\left(x_t,p,s,g\right)$. Finally, the sampled latent code $x_0$ is decoded into video pixels $v$ by image autoencoder.

\section{Experiments}

\subsection{Datasets}
In the training process, we utilize WebVid and VideoHD to optimize DragNUWA. 
\begin{itemize}
\item \textbf{WebVid} is a vast dataset~\citet{bainFrozenTimeJoint2021} comprising 10 million web videos encompassing diverse real-world scenarios with corresponding caption. It covers a wide range of motion patterns, making it suitable for open-domain trajectory-based video generation.

\item \textbf{VideoHD} We build VideoHD dataset based on web-crawled videos. We first collected 75K high-resolution, top-quality video clips from the internet. Subsequently, these clips are annotated using BLIP2~\citet{liBLIP2BootstrappingLanguageImage2023}. Finally, we manually filter out some errors in the generated results.

\end{itemize}

\subsection{Implementation Details}
\begin{table}[]
\renewcommand{\arraystretch}{1.1}
\begin{tabular}{|p{0.65cm}p{3.05cm}|p{4.0cm} p{4.0cm}|}
\hline
\multicolumn{2}{|c|}{Version}                                                           & \multicolumn{1}{c|}{DragNUWA-LD}                                                                                                             & \multicolumn{1}{c|}{DragNUWA-HD}                                                                                                                    \\ \hline
\multicolumn{1}{|c|}{\multirow{6}{*}{Data}}                    & \multicolumn{1}{c|}{Dataset}                & \multicolumn{1}{c|}{WebVid}                                                                                                   & \multicolumn{1}{c|}{WebVid+VideoHD}                                                                                       \\ \cline{2-4} 
\multicolumn{1}{|c|}{}                                         & \multicolumn{1}{c|}{Samples}                & \multicolumn{1}{c|}{10M}                                                                                                            & \multicolumn{1}{c|}{10M+75K}                                                                                                             \\ \cline{2-4} 
\multicolumn{1}{|c|}{}                                         & \multicolumn{1}{c|}{Resolution $(W\times H)$}       & \multicolumn{1}{c|}{$320\times 192$}                                                                                                        & \multicolumn{1}{c|}{$576\times 320$}                                                                                                             \\ \cline{2-4} 
\multicolumn{1}{|c|}{}                                         & \multicolumn{1}{c|}{Max Duration}      & \multicolumn{1}{c|}{2s}                                                                                                             & \multicolumn{1}{c|}{4s}                                                                                                                  \\ \cline{2-4} 
\multicolumn{1}{|c|}{}                                         & \multicolumn{1}{c|}{Frames $(L)$}             & \multicolumn{1}{c|}{8f}                                                                                                             & \multicolumn{1}{c|}{16f}                                                                                                                 \\ \cline{2-4} 
\multicolumn{1}{|c|}{}                                         & \multicolumn{1}{c|}{Framerate}              & \multicolumn{2}{c|}{4 fps}                                                                                                                                                                                                                                \\ \hline
\multicolumn{1}{|c|}{\multirow{3}{*}{Augmentation}}            & \multicolumn{1}{c|}{RandomResizeCrop}       & \multicolumn{2}{c|}{scale=(0.9, 1.), ratio=(5/3, 5/3)}                                                                                                                                                                                                    \\ \cline{2-4} 
\multicolumn{1}{|c|}{}                                         & \multicolumn{1}{c|}{ColorJitter}            & \multicolumn{2}{c|}{brightness=0.05, contrast=0.15, saturation=0.15}                                                                                                                                                                                        \\ \cline{2-4} 
\multicolumn{1}{|c|}{}                                         & \multicolumn{1}{c|}{RandomStartFrame}       & \multicolumn{1}{c|}{[0, video\_duration-2]}                                                                                     & \multicolumn{1}{c|}{[0, video\_duration-4]}                                                                                          \\ \hline
\multicolumn{1}{|c|}{\multirow{3}{*}{\shortstack{Trajectory \\ Sampler(TS)}}} & Max Trajectories   $(N)$  & \multicolumn{2}{c|}{8}                                                                                                                                                                                                                                    \\ \cline{2-4} 
\multicolumn{1}{|c|}{}                                         & \multicolumn{1}{c|}{Gaussian Kernel}        & \multicolumn{2}{c|}{kernel\_size=99,   sigma=10}                                                                                                                                                                                                          \\ \cline{2-4} 
\multicolumn{1}{|c|}{}                                         & \multicolumn{1}{c|}{Anchor Interval ($\lambda$)}    & \multicolumn{2}{c|}{16}                                                                                                                                                                                                                                   \\ \hline
\multicolumn{1}{|c|}{\multirow{4}{*}[-8ex]{\shortstack{Multiscale \\ Fusion(MF)}}}  & \multicolumn{1}{c|}{Text Control $(p)$}       & \multicolumn{2}{c|}{77x1024}                                                                                                                                                                                                                             \\ \cline{2-4} 
\multicolumn{1}{|c|}{}                                         & \multicolumn{1}{c|}{Image Control $(s)$}      & \multicolumn{1}{c|}{\begin{tabular}[c]{@{}c@{}}$8\times 320\times 40\times 24$\\  $8\times 320\times 20\times 12$\\ $8\times 640\times 10\times 6$\\   $8\times 1280\times 5\times 3$ \end{tabular}} & \multicolumn{1}{c|}{\begin{tabular}[c]{@{}c@{}} $16\times 320\times 72\times 40$\\ $16\times 320\times 36\times 20$\\   $16\times 640\times 18\times 10$\\  $16\times 1280\times 9\times 5$ \end{tabular}} \\ \cline{2-4} 
\multicolumn{1}{|c|}{}                                         & \multicolumn{1}{c|}{Trajectory Control $(g)$} & \multicolumn{1}{c|}{\begin{tabular}[c]{@{}c@{}}$8\times 320\times 40\times 24$\\  $8\times 320\times 20\times 12$\\   $8\times 640\times 10\times 6$\\  $8\times 1280\times 5\times 3$\end{tabular}} & \multicolumn{1}{c|}{\begin{tabular}[c]{@{}c@{}}$16\times 320\times 72\times 40$\\  $16\times 320\times 36\times 20$\\   $16\times 640\times 18\times 10$\\  $16\times 1280\times 9\times 5$\end{tabular}} \\ \cline{2-4} 
\multicolumn{1}{|c|}{}                                         & \multicolumn{1}{c|}{Control Drop Ratio}     & \multicolumn{2}{c|}{Text: 0.1, Image: 0.1, Trajectory: 0.1}                                                                                                                                                                                             \\ \hline
\multicolumn{1}{|c|}{\multirow{5}{*}{\shortstack{Adaptive \\ Training(AT)}}}  & \multicolumn{1}{c|}{Batch Size}              & \multicolumn{2}{c|}{128}                                                                                                                                                                                                                                  \\ \cline{2-4} 
\multicolumn{1}{|c|}{}                                         & \multicolumn{1}{c|}{Learning Rate}           & \multicolumn{2}{c|}{$5\times 10^{-6}$}                                                                                                                                                                                                                               \\ \cline{2-4} 
\multicolumn{1}{|c|}{}                                         & \multicolumn{1}{c|}{Scheduler}              & \multicolumn{2}{c|}{WarmupLinear, warmup\_ratio=0.05}                                                                                                                                                                                                           \\ \cline{2-4} 
\multicolumn{1}{|c|}{}                                         & \multicolumn{1}{c|}{Optimizer}              & \multicolumn{2}{c|}{Adam}                                                                                                                                                                                                                                 \\ \cline{2-4} 
\multicolumn{1}{|c|}{}                                         & \multicolumn{1}{c|}{Parameters}             & \multicolumn{2}{c|}{1.60B}                                                                                                                                                                                                                                \\ \hline
\end{tabular}
\caption{Implementation details of DragNUWA.}
\label{tab:implementation_details}
\end{table}

We implement two versions of DragNUWA, namely DragNUWA-LD and DragNUWA-HD. DragNUWA-LD is trained on videos of 8 frames with a resolution of $320\times 192$, while DragNUWA-HD is trained on 16 frames with a resolution of $576\times 320$. For the Trajectory Sampler (TS), the maximum number of trajectories $N$ is 8, with anchor interval $\lambda$ of 16. The Gaussian kernel size is 99, with sigma value set to 10. To support different condition combinations, we randomly omit text, images, and trajectories with a probability of 0.1. We train the model using Adam optimizer~\citet{kingmaAdamMethodStochastic2014} with a batch size of $128$, learning rate of $5\times 10^{-6}$. More implementation details can be found in Tab.~\ref{tab:implementation_details}.

\subsection{Trajectory Controllability}
Contrary to existing studies that focus on text or image control, DragNUWA primarily emphasizes modeling trajectory control. In order to validate the effectiveness of trajectory control, we test DragNUWA from two aspects: camera movements and complex trajectories.

\begin{figure}[h]
    \makebox[\linewidth]{
        \includegraphics[width=\linewidth]{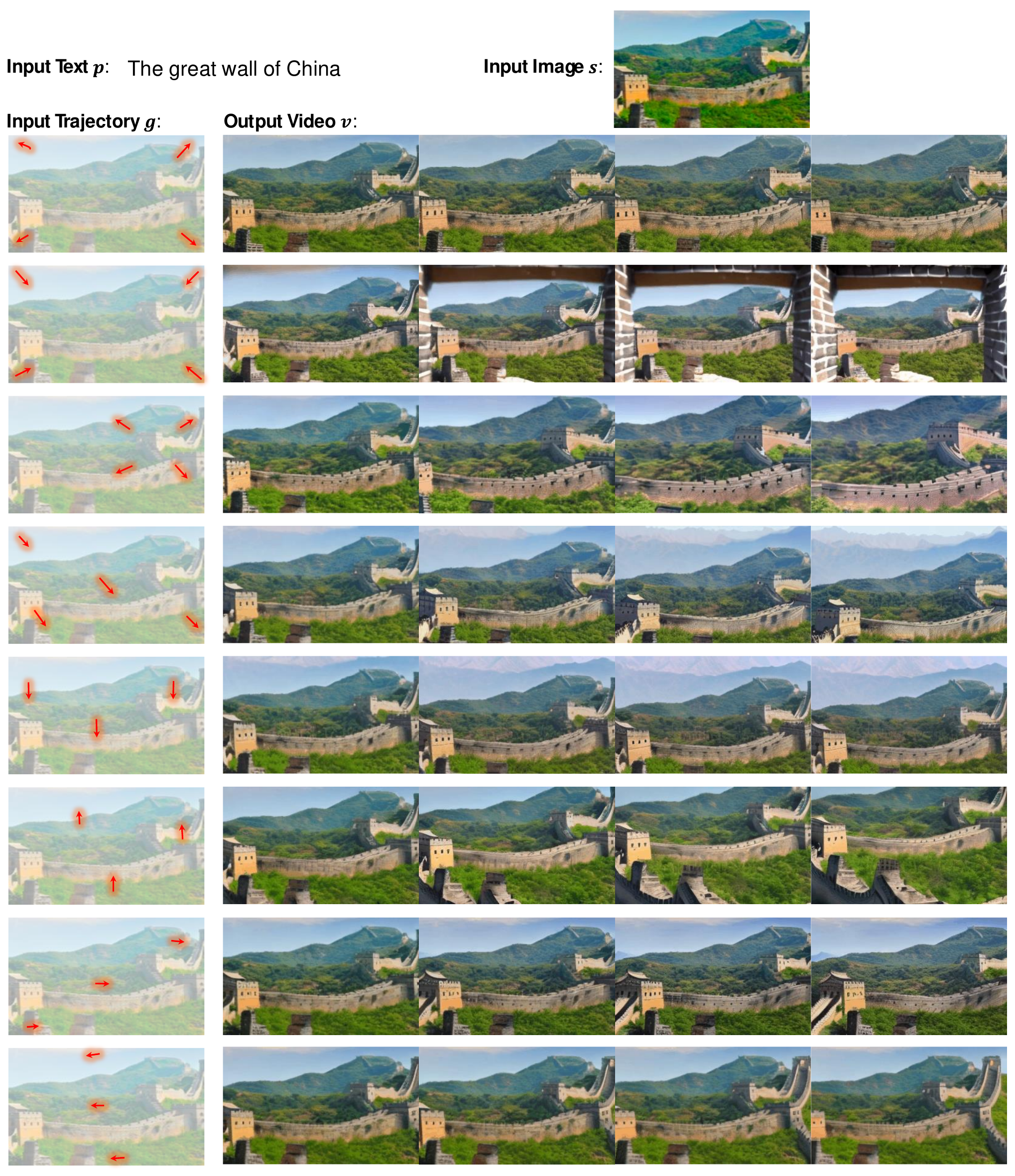}
    }
    \caption{Various camera movement effects can be achieved by utilizing identical text and images while altering  the dragging trajectories. For instance, zoom-in and zoom-out effects can be expressed by drawing the directional trajectories at the desired zoom locations.}
    \label{fig:camera}
\end{figure}

\textbf{Camera movements.} In video production, camera movements play a significant role in creating dynamic and engaging visuals for the audience. Different types of camera movements can aid in narrative storytelling, or emphasizing elements within a scene. Common camera movements include not only horizontal and vertical movements but also zooming in and zooming out. As shown in Fig.~\ref{fig:camera}, we discovered that although DragNUWA does not explicitly model camera movements, it  learns various camera movements from the modeling of open-domain trajectories.

\begin{figure}[h]
    \makebox[\linewidth]{
        \includegraphics[width=\linewidth]{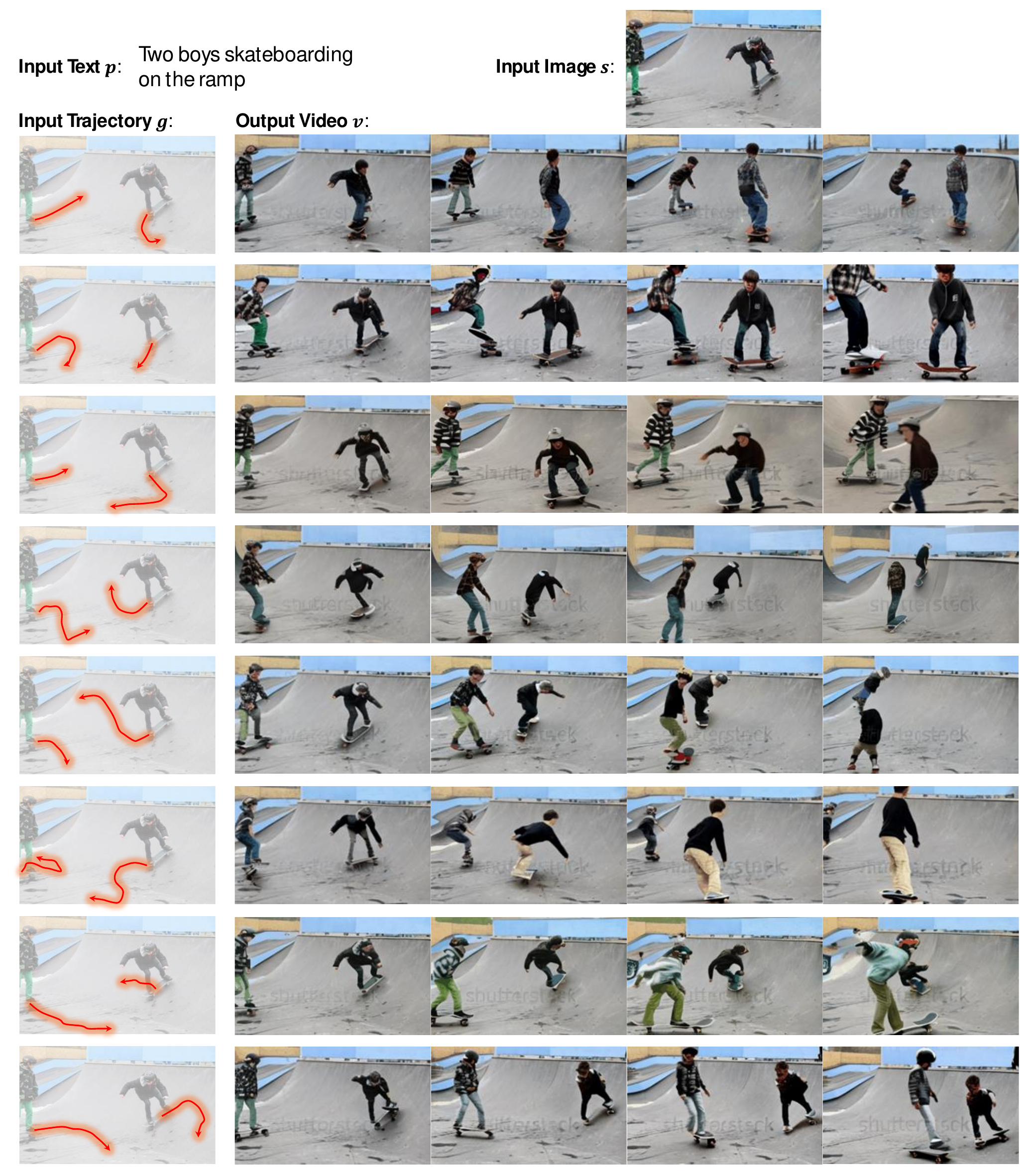}
    }
    \caption{Various complex trajectory effects can be achieved by employing the same text and image while altering the dragging trajectory. DragNUWA supports complex curved trajectories, allows for variable trajectory lengths, and supports concurrent control of trajectories for multiple objects.}
    \label{fig:complex}
\end{figure}

\textbf{Complex Trajectories.} Motion modeling in video generation presents challenges due to the presence of multiple moving objects, intricate motion trajectories, and varying motion amplitudes among different objects. To evaluate the capability of DragNUWA in accurately modeling complex motion, we conducted tests on various intricate drag trajectories using the same image and text, as depicted in Fig.~\ref{fig:complex}. Our findings indicate that DragNUWA can reliably control complex motions. This encompasses several aspects: firstly, DragNUWA supports complex curved trajectories, enabling the generation of objects moving along the specific intricate trajectory (see Row 6). Secondly, DragNUWA allows for variable trajectory lengths, with longer trajectories resulting in larger motion amplitudes (see Row 7-8). Lastly, DragNUWA has the capability to simultaneously control the trajectories of multiple objects. To the best of our knowledge, no existing video generation model has effectively achieved such trajectory controllability, highlighting DragNUWA's substantial potential to advance controllable video generation in future applications.

\subsection{Essential of three controls}

\begin{figure}[h]
    \makebox[\linewidth]{
        \includegraphics[width=\linewidth]{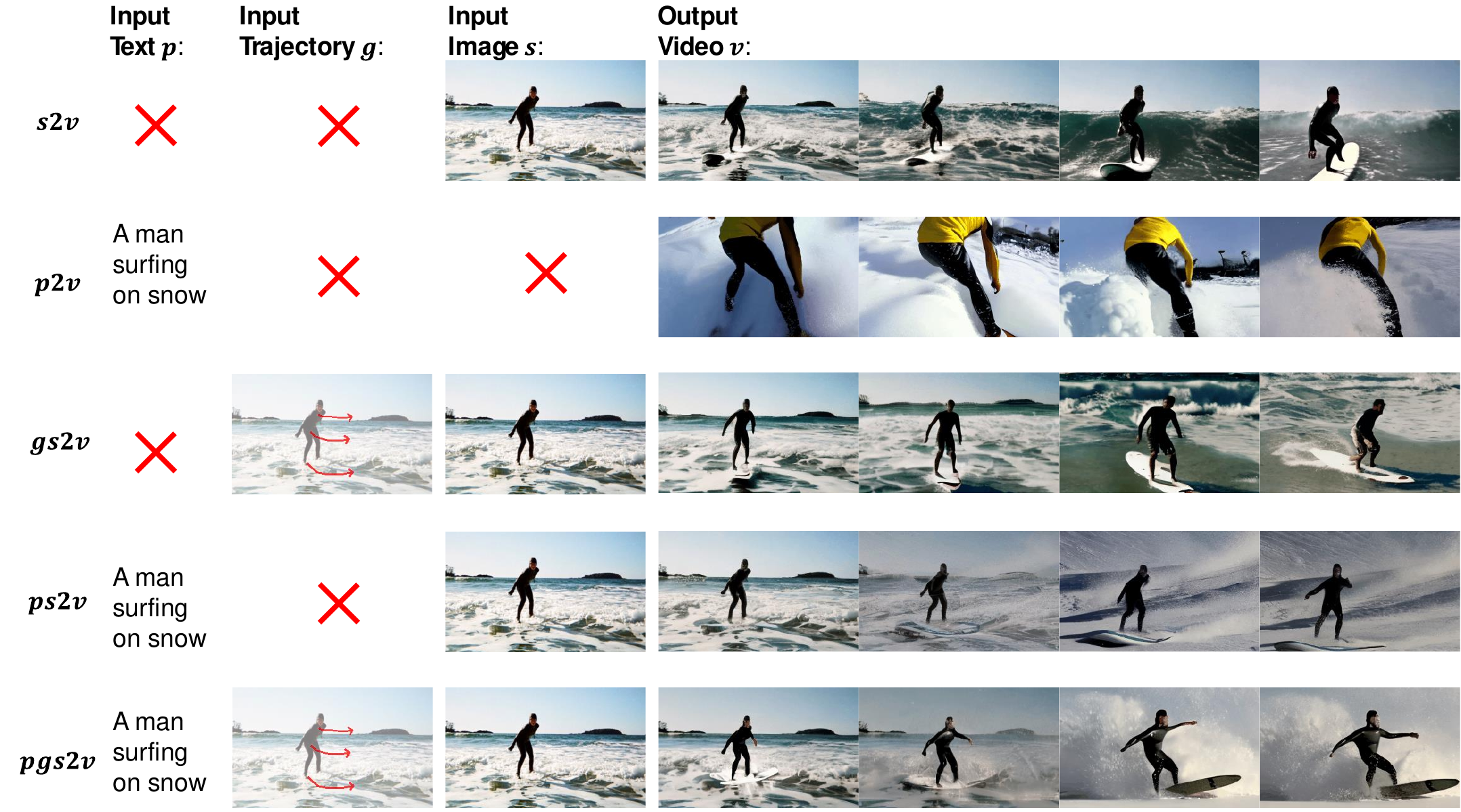}
    }
    \caption{DragNUWA achieves  fine-grained video generation by integrating three essential controls: text, image, and trajectory, corresponding to semantic, spatial, and temporal aspects, respectively.}
    \label{fig:three}
\end{figure}

While DragNUWA primarily emphasizes trajectory control modeling, it also incorporates the control of text and images. We believe that text, image, and trajectory each represent one of the three fundamental control aspects of videos: semantic, spatial, and temporal perspectives. Fig.~\ref{fig:three} illustrates  the necessity of these conditions by showcasing different combinations of text $(p)$, trajectory $(g)$, and image $(s)$, including $s2v$, $p2v$, $gs2v$, $ps2v$, and $pgs2v$. It is important to note that we did not model $g2v$ and $pg2v$, as we believe that trajectories without images are meaningless.

The $s2v$ and $p2v$ exemplify the constraints of image and text control when utilized as an individual condition. As shown in $s2v$, although an image alone provides some potential semantic and kinetic information, it does not allow for precise control over the background and the character's movement. As illustrated in $p2v$,  when only text is provided, the model successfully generates a video related to the text, however, the appearance and dynamics remain entirely uncontrollable. The $gs2v$ and $ps2v$ emphasize the importance of text $(p)$ and trajectory $(g)$. In the absence of text, it is impossible to determine whether the ambiguous image $(s)$ represents surfing on the sea or snow. In the absence of trajectory, the model automatically assumes that the character is moving to the left. The $pgs2v$ demonstrates the combined impact of all three essential conditions, enabling the control of surfing on the snow and moving to the right.

It is worth mentioning that some studies incorporate video as a condition, which is beyond the scope of this research. We focus on the fundamental conditions, while the video condition provides excessive information, significantly constraining the creation of videos and primarily serving for style transfer purposes. Moreover, the video condition necessitates users to provide specific video materials, consequently presenting significant challenges in practical application.

\section{Conclusion}
We present DragNUWA, an end-to-end video generation model that seamlessly incorporates text, image, and trajectory input, enabling fine-grained and user-friendly control from semantic, spatial, and temporal perspectives. Additionally, our trajectory modeling framework, consisting of the Trajectory Sampler (TS), Multiscale Fusion (MF), and Adaptive Training (AT), tackles challenges in open-domain trajectory control, thereby enabling the generation of coherent videos in accordance with complex trajectories. Experiments validate DragNUWA's superiority over existing approaches, demonstrating its ability to generate fine-grained videos effectively.

\bibliography{iclr2024_conference}
\bibliographystyle{iclr2024_conference}

\end{document}